\documentclass[conference]{IEEEtran}
\IEEEoverridecommandlockouts
\usepackage{cite}
\usepackage{amsmath,amssymb,amsfonts}
\usepackage{algorithmic}
\usepackage{graphicx}
\usepackage{textcomp}
\usepackage{xcolor}
\usepackage{enumitem}

\def\BibTeX{{\rm B\kern-.05em{\sc i\kern-.025em b}\kern-.08em
    T\kern-.1667em\lower.7ex\hbox{E}\kern-.125emX}}
\begin{document}

\title{Service-Based Drone Delivery\\}

\author{
\IEEEauthorblockN{Balsam Alkouz}
\IEEEauthorblockA{\textit{School of Computer Science} \\
\textit{The University of Sydney}\\
Australia \\
balsam.alkouz@sydney.edu.au}

\and
\IEEEauthorblockN{Babar Shahzaad}
\IEEEauthorblockA{\textit{School of Computer Science} \\
\textit{The University of Sydney}\\
 Australia \\
babar.shahzaad@sydney.edu.au}

\and
\IEEEauthorblockN{Athman Bouguettaya}
\IEEEauthorblockA{\textit{School of Computer Science} \\
\textit{The University of Sydney}\\
Australia \\
athman.bouguettaya@sydney.edu.au}

}

\maketitle

\begin{abstract}
Service delivery is set to experience a major paradigm shift with fast advances in drone technologies coupled with higher expectations from customers and increased competition. We propose a novel service-oriented approach to enable the ubiquitous delivery of packages in a drone-operated \textit{skyway network}. We discuss the benefits, framework and architecture, contemporary approaches, open challenges and future visioned directions of service-based drone deliveries.
\end{abstract}


\section{Introduction}
Smart cities are composed of several smart components including smart homes, smart agriculture, smart buildings, smart campus, smart economy, smart logistics, etc. \cite{9086495}. Smart cities integrate various technologies that offer innovative services to improve the quality of citizens' life \cite{Mishra2020}. The key stakeholders of smart cities include citizens, governments, urban businesses, and service providers \cite{jayasena2019stakeholder}. In this respect, Unmanned Aerial Vehicles (UAVs) offer unique capabilities and functionalities required to realize the vision of smart cities \cite{MOHAMED2020119293}. A UAV is a flying robot that is controlled remotely or is operated autonomously to perform specific tasks \cite{valavanis2015handbook}. A drone is a specific type of UAV that provides support for a variety of civilian applications in smart cities \cite{DBLP:journals/corr/abs-1805-00881}. Examples of drone applications include traffic monitoring, firefighting, surveillance, agriculture, and package delivery \cite{2}. Drones provide a ubiquitous (i.e., anytime anywhere), cost-effective, fast, contactless, and environmentally-friendly alternative for package delivery to end users. Several logistics companies such as Amazon, Google, Alibaba are stepping up efforts to use drones for delivery services \cite{Aurambout2019}.

The \textit{service paradigm} provides a key mechanism to abstract the \textit{functional} capabilities of drones and their \textit{non-functional} attributes as drone delivery services \cite{Bouguettaya:2017:SCM:3069398.2983528,shahzaad2019composing}. In this context, the key functional aspect of a drone delivery service is the {\em transport of a package} from one location (e.g., warehouse rooftop) to another location (e.g., customer's building rooftop) in a {\em skyway} network \cite{10.1145/3460418.3479289}. The non-functional (i.e., \textit{Quality of Service} (QoS)) aspect of a drone delivery service may include payload capacity, flight range, battery capacity, etc. A skyway network is made up of a set of line segments whose endpoints constitute the nodes of the network \cite{9284115}. These nodes typically represent the rooftops of buildings and dwellings. The nodes may concurrently act as both \textit{delivery targets} and/or \textit{recharging stations}. The transport of a package along a line segment would represent a \textit{drone delivery service} operating under a set of {\em constraints}.

Drones for delivery services present unique challenges to fully deliver on their potential. In this respect, the \textit{inherent} limitations and \textit{contextual} constraints of a drone exhibit key challenges for the wide-range deployment of drone-based delivery services. Examples of inherent limitations include limited payload, limited flight range, and battery capacity \cite{jermaine2021demo}. Examples of contextual constraints include those related to recharging pads at each recharging station as well as uncertain weather conditions such as wind \cite{alkouz2021provider}. In addition, current drone flying regulations only allow the use of small drones for delivery (payload $<$ 2.5 kg). There are instances where there is a need to deliver goods by a deadline and which weigh more than the maximum of a single drone’s payload. In this case, the use of \textit{drone swarms} is an effective alternative to address the aforementioned regulations and requirements \cite{alkouz2020swarm}. However, the use of drone swarms also presents unique challenges when used for delivery services. These challenges are mostly related to the requirement that all drones in a swarm arrive together within a time window, their varying flight performance, formation, etc. \cite{alkouz2020formation}.

\begin{figure} [tp!]
    \centering
    \includegraphics[width=0.9\linewidth]{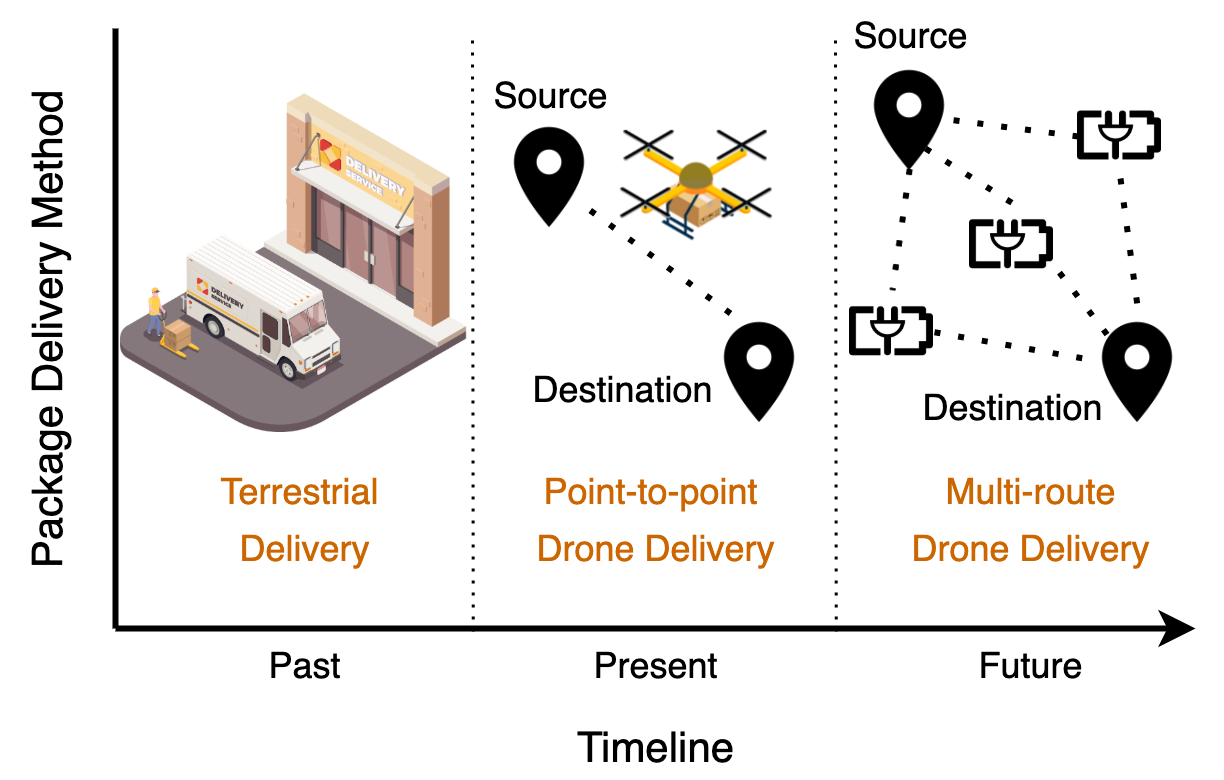}
    \caption{The Evolution of Package Delivery Methods}
    \label{pastpresentfuture}
\end{figure}

\begin{figure} [t]
    \centering
    \includegraphics[width=\linewidth]{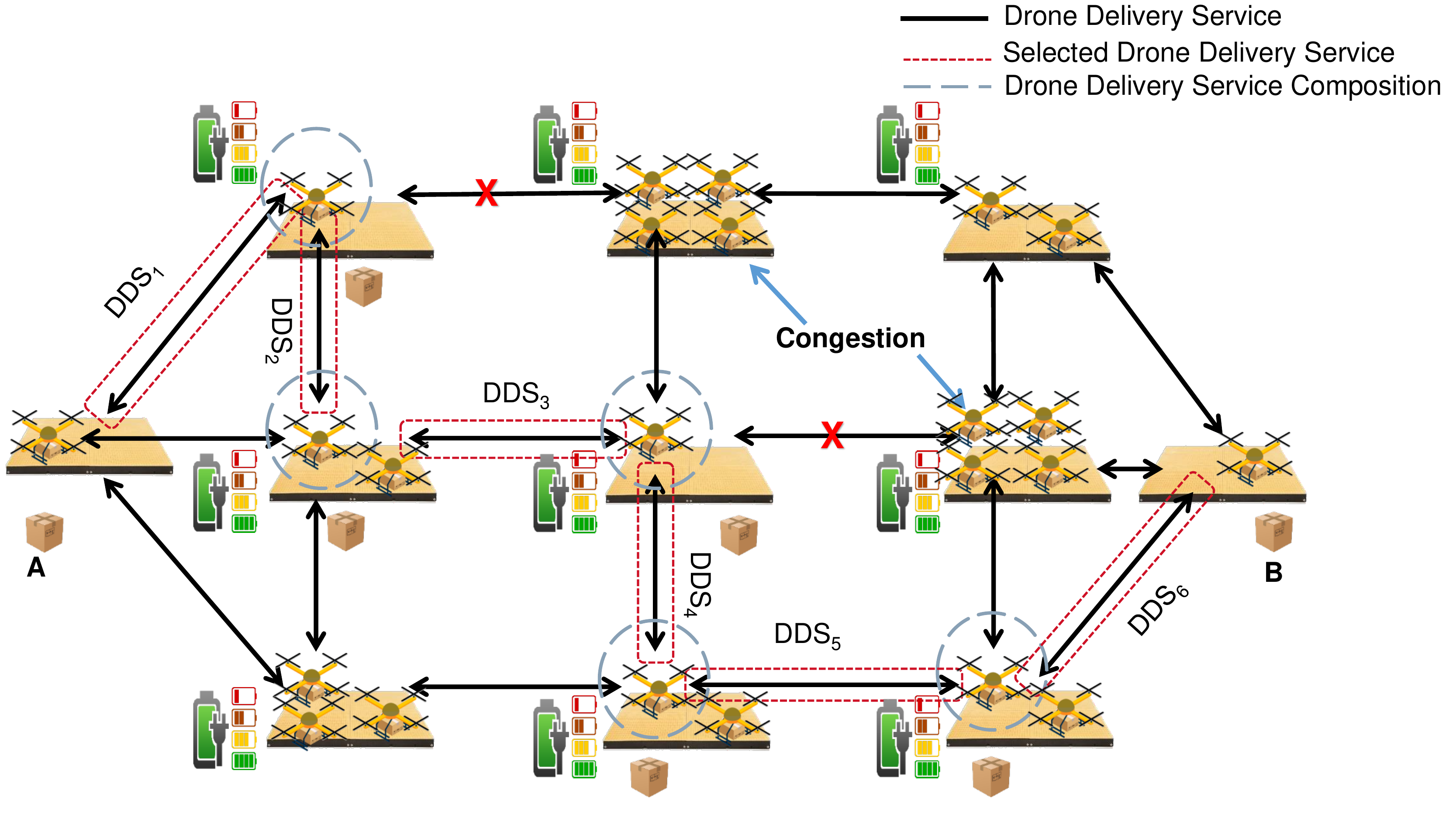}
    \caption{Drone Delivery Service Composition in a Skyway Network}
    \label{fig2}
\end{figure}

The evolution of package delivery methods is presented in Fig. \ref{pastpresentfuture}. Current approaches mainly focus on point-to-point drone-based deliveries ignoring the drone’s recharging requirements and complexity of the environments \cite{7513397}. In these approaches, the drone delivery problem is usually formulated as \textit{Travelling Salesman Problem} (TSP) \cite{8488559} or \textit{Vehicle Routing Problem} (VRP) \cite{WANG2019350}. However, there is an increasing number of studies that focus on service-based approaches for multi-point single drone-based and drone swarm-based deliveries in skyway networks \cite{alkouz2020swarm}. These service-based approaches take into account different constraints affecting the delivery services, e.g., payload capacity, battery capacity, wind conditions, etc. A single drone delivery service may not satisfy a delivery plan because of the aforementioned challenges. In such cases, a service {\em composition} is required to deliver packages. An \textit{optimal service composition} is defined as the selection and aggregation of the {\em best} drone delivery services (i.e., skyway segments) in a skyway network from a given source to a destination \cite{2021335}. The composition of drone delivery services creates a {\em value-added} service while meeting the QoS requirements of end-users which include fast, safe, cost-effective, and contactless delivery. Fig. \ref{fig2} presents a drone delivery service composition scenario where a drone delivers a package from point A to point B. In this scenario, a drone may not travel directly from the source to the destination due to flight regulations and flight range limitations. As each drone service has its specific QoS attributes such as flight time and delivery cost. Therefore, different compositions will provide different aggregate QoS.

This paper maps out a strategy to leverage the service paradigm for the potential utilization of drones for delivery services in smart cities. We envision a green and sustainable logistics future for smart cities provided by drone delivery services. We highlight the benefits of service-based drone delivery from providers and consumers perspective. We present a service-based drone delivery framework that includes enabling technologies, operating environment, types of deliveries, and service-oriented architecture of sensor-cloud infrastructure. In this paper, we also provide the details of contemporary approaches for service-based drone delivery. Finally, we describe the open challenges and draw a road map for future research directions.\looseness=-1


\section{Drone Delivery Services: A Sustainable Logistics Future}
Drone-based delivery services provide a myriad of benefits from dual perspectives: service providers and service consumers perspectives. Hereinafter, we describe the benefits from each perspective in detail.
\begin{itemize}[leftmargin=*]
    \item \textbf{Service Provider Perspective:} A drone delivery service provider is typically an owner and operator of a fleet of drones that manages the delivery of packages to end consumers \cite{alkouz2021provider}. A provider may be a warehouse that owns its own fleet to deliver its products. A provider may also be a shipping and logistics company like Amazon, Wing, and UPS. Here, we discuss some of the main benefits from a service provider perspective.
    \begin{itemize}
    \item \textit{Labor efficiency:} A service provider mainly benefits from cutting costs and increasing profit. By using drones, delivery companies drastically cut down the cost of human labor \cite{aurambout2019last}. Unlike ground deliveries that require many drivers, one human operator can manage a fleet of drones. Human involvement will be required to monitor flights and perform maintenance tasks only. One employee would typically be able to monitor multiple flights at once similar to an air-traffic controller in commercial passenger planes \cite{sudbury2016cost}. It is evaluated that the drone cost per package delivery is one-third or less of the UPS ground delivery cost \cite{sudbury2016cost}.
        
    \item \textit{Growing industry:} A recent study estimates that drones could account for one-third of same-day deliveries by 2030 \cite{market}. This growth will reflect on the profit of delivery service providers who are adopting the drone technology early \cite{alkouz2021provider}. This rate has recently seen a boost since the start of the COVID-19 pandemic. Expansive testing and adoption of drone delivery systems was accelerated due to lockdowns, quieter skies, and need for contactless deliveries \cite{shahzaad2021robust}. The drone delivery service market is forecasted to grow at a significant compound annual growth rate of 14.5\% between 2023 to 2030 \cite{market}.

    \item \textit{Nature of delivery orders:} According to Amazon, 86\% of their deliveries are under 5-pound weight \cite{pierce2013delivery}. This weight limit is what most commercial drones could carry and what flight authorities regulations allow\footnote{https://www.faa.gov/uas/advanced\_operations/package\_delivery\_drone}. Hence, delivery service providers adopting drones can rest assure that their drones will be highly demanded. In addition, due to the fast delivery options offered by many businesses (e.g. priority delivery in Uber Eats), a drone delivery service provider may charge extra for this added value that it can offer without any added cost \cite{sudbury2016cost}.

    \item \textit{Medium and routing efficiency:} Air travel is the fastest method to get from point A to B. This is due to the medium that has less friction compared to the friction associated with ground travel. Therefore, resulting in faster deliveries that allow the reutilization of drones multiple times a day \cite{alkouz2021provider}. In addition, the route nature of air travel, whether it is point-to-point or managed traffic in skyways, make it easier to allocate resources and compute near precise delivery times and costs \cite{shahzaad2019constraint}. Such factors are unavoidable in ground deliveries due to the infrastructure and unpreventable traffic. For example, in ground delivery, there exists many instances where a multi-destination delivery trip is planned in an inefficient order \cite{wang2019vehicle}. Such problems are mitigated in air which enhances reach and allows hard-to-reach parts to be catered \cite{shahzaad2019composing}. Drone delivery has the ability to overcome difficult ground terrains and rivers due to the privilege of using the air medium \cite{YOO20181687}.


    \item \textit{Carbon gains:} In addition to quantitative gains, drone deliveries bring down carbon emissions \cite{goodchild2018delivery}. This is essential for an environmentally friendly branding for a business. Moreover, an electric source of power that drones require typically cost less than fuel-based ground deliveries \cite{chiang2019impact}.\looseness=-1
\end{itemize}
    
    \item \textbf{Service Consumer Perspective:} A service consumer may be defined as the end customer who orders the packages. Alternatively, a drone-based service consumer may be called on warehouses, restaurants, and other beneficiaries who may utilize such services for the efficient shipment of their products. Here, we discuss some of the main benefits form a service consumer perspective.
    \begin{itemize}
        \item \textit{Fast delivery:} Drones offer a faster delivery option of packages compared to terrestrial deliveries \cite{yoo2018drone}. Alphabet's Wing flies at a speed of 113km/h ensuring the delivery of packages within minutes\footnote{https://wing.com/en\_au/}. Another example is Amazon Prime Air, their service aims to drop packages in less than 30 minutes \cite{sudbury2016cost}. This feature is especially significant for the delivery of emergency parcels, e.g. medicine and defibrillators \cite{mermiri2020drones}.
        
        \item \textit{Cost effective:} Using electric and autonomous drones for delivery can result in lower costs \cite{chiang2019impact}. These cost savings can be passed down to the end consumers through a reduction in service prices. This cost reduction is mainly due to the cheap drone technology, and the ability to cover greater distances in shorter times, thereby saving fuel and cutting costs per mile \cite{chiang2019impact}. It is estimated that drones' operational cost are at least 70\% less than van delivery services \cite{goasduff2020flying}.
        \item \textit{Convenience and reach:} With the increase in consumer's desire for speed, on-demand drone delivery can help in achieving consumers' convenience \cite{yoo2018study}. Drone deliveries have the ability to overcome road-traffic and deliver fast \cite{yoo2018drone}. Furthermore, due to the lack of infrastructure, complicated terrain, and extreme weather conditions, delivery of emergency supplies to remote areas are often hampered with typical means of ground deliveries. Therefore, drone deliveries have proved critical in getting goods to the last mile and overcoming commuting, delivery, and distribution challenges \cite{salama2020joint}.
        
        \item\textit{Safer delivery:} The contactless drone delivery nature offers a safer alternative to human involved deliveries. Especially at times of pandemics, the need for contactless deliveries increase to reduce the spread of diseases \cite{alkouz2021provider}. Moreover, the contactless nature may reduce the probability of assault attacks by fake delivery drivers ensuring the safety of consumers \cite{bbc_news_2021}. Furthermore, many people developed face-to-face interactions social anxieties whose causes would be mitigated due to the contactless nature of drone deliveries \cite{jefferies2020social}. \looseness=-1
      
    \end{itemize}
\end{itemize}



\section{Service-Based Drone Delivery Framework}
A service-based drone delivery framework composes of four main components: (1) the drones as the enabling technology of the service, (2) the skyway network as the operating environment, (3) the delivery mission types, and (4) the service-oriented architecture of sensor-cloud infrastructure as the operation facilitator.

\subsection{Enabling Technology}
Drones are an enabling technology for service-based drone deliveries in smart cities. Delivery drones are generally categorized into one of three types based on their fabrication. Each of these types has its own pros and cons in terms of payload, flexibility, speed, flight range, battery, and takeoff and landing requirements.

\begin{itemize}[leftmargin=*]
    \item \textbf{Multi-Rotor Drones:} Multi-rotor drones are the most popular type of delivery drones \cite{yang2017multi}. The most common uses of multi-rotor drones include aerial spraying, crowd monitoring, photography, and package delivery. These drones are preferred where hovering and vertical take-off and landing is required. In addition, the multi-rotor drones provide the ability to maneuver through tight spaces, e.g., high-rise buildings in urban areas. However, these drones have shorter flight times compared to fixed-wing drones. A multi-rotor drone with the payload weight typically stays in the air up to 25 minutes \cite{10.1007/978-3-319-43506-0_44}.
    
    \item \textbf{Fixed-Wing Drones:} Fixed-wing drones have similar configurations as passenger airplanes \cite{10.1007/978-981-33-6981-8_49}. These drones can carry larger payload weights and travel long-range distances at a fast speed. A fixed-wing drone can typically fly up to 45 minutes. The main limitation of a fixed-wing drone is its inability to hover in one place. In addition, it requires a dedicated runway to take-off and land.
    \item \textbf{Hybrid Drones:} Hybrid drones are a special type of drones that offer a combination of vertical take-off and landing from a limited space, hovering, and flight capabilities including long-range and high-speed \cite{SAEED201891}. Hybrid drones are still in the early development stage \cite{goetzendorf2021lightweight}. However, they are expected to dominate both military and civilian applications. Amazon has tested a hybrid drone which has both multi-rotor and fixed-wing characteristics to deliver packages \cite{7802742}.
    
\end{itemize}

\subsection{Operating Environment}
The drone delivery services operate in skyway networks for multi-point package deliveries \cite{9284115}. A skyway network enables the safe and scalable deployment of drone-based delivery solutions in a shared airspace. A skyway network links a set of predefined line segments between any two nodes to provide conflict-free routes to drones (Fig. \ref{fig2}). In this context, a conflict-free route follows the line-of-sight drone flying regulations and avoids flying over no-fly zones. We assume that a skyway network contains virtual barriers to prevent drones’ access to restricted areas such as airports and other sensitive sites. Each node in the skyway network is a fixed landing pad on a rooftop of a building. A single drone may not be able to travel for long distances without recharging as most drones fly for an average of only 30 minutes before requiring a battery recharge \cite{abdallah2019efficient}. Therefore, each node in the skyway network may concurrently act as a recharging station. We use existing infrastructure of a city where each building rooftop may be easily and cheaply fitted with a wireless recharging pad. The vision of having a skyway network infrastructure is both feasible and realistic to enable drone-based delivery services. Indeed, the building rooftops infrastructure already exists enabling the deployment of recharging stations at a minimum cost.

\subsection{Types of Delivery}

\subsubsection{Single Drone Delivery}
This section describes the utilization of a single drone for the provisioning of delivery services. In many instances, a single drone can satisfy the delivery requirements of customers. Delivery drones can carry packages weighing less than five pounds (2.27 kg) and cover about 86 percent of Amazon's items \cite{doi:10.1111/drev.10313}. However, there are a number of operational constraints that hinder the potential deployment of single drone delivery services. The operational constraints for drone delivery services include the drone's flight range, weather conditions, and availability of pads at recharging stations. The range of a typical delivery drone with full payload weight varies from 3 to 33 km \cite{12}. Therefore, a drone may need multiple times of recharging to serve long-distance areas.

Service-based approaches have been proposed in recent years to address the aforementioned operational constraints in context of a single drone-based delivery \cite{babar2021topk,9284115}. It is quite natural to model the drone delivery using the service paradigm because it maps to the key ingredients, i.e., functional and non-functional attributes. The drone delivery services usually operate in skyway networks where the composite services provide value-added benefits to the package delivery using drones. In this respect, a composite service represents an aggregation of a set of best drone services meeting the customer's expectations in terms of delivery time and cost. A constraint-aware drone service composition approach ensures the selection and composition of drone services that avoid the congestion conditions at recharging stations \cite{shahzaad2019constraint}. A robust service composition is proposed taking into account the uncertain weather conditions such as wind \cite{shahzaad2021robust}. To address the failures in drone services, a resilient drone service composition exists that is built upon the constraint-aware service composition \cite{2021335}.

\subsubsection{Swarm-based Drone Delivery}
The use of multiple drones cooperatively to accomplish a common objective is a natural progression of using single drones in various applications, including delivery. Drone swarms represent a collection of UAVs capable of synchronizing, coordinating, and communicating with each other to accomplish a task as one cohesive unit \cite{tahir2019swarms}. In delivery, there may be instances where the delivery of multiple/heavier packages is hampered by the flight regulations that only allow the operation of small drones (payload $<$ 2.5kg) in a city\footnote{https://www.faa.gov/uas/advanced\_operations/package\_delivery\_drone}. Heavier/multiple items necessitate larger drones which can be costly and undesirable to the general public and regulators due to a variety of reasons \cite{alkouz2021provider}. The noise that huge drones produce, as well as the fact that the impact of a mistake is likely to be greater for large drones than for small drones, are among these issues. Expanding drones' capacity to heavier/multiple items could revolutionise the logistics industry, therefore there's a lot of interest in overcoming the payload restriction. Swarm technology for delivery is one of the most promising of the options being investigated \cite{alkouz2020swarm}. Furthermore, swarms of drones in delivery are capable of covering longer distances as the payload-battery pressure may be distributed amongst several drones \cite{alkouz2020formation}.

\begin{figure} [htp!]
    \centering
    \includegraphics[width=\linewidth]{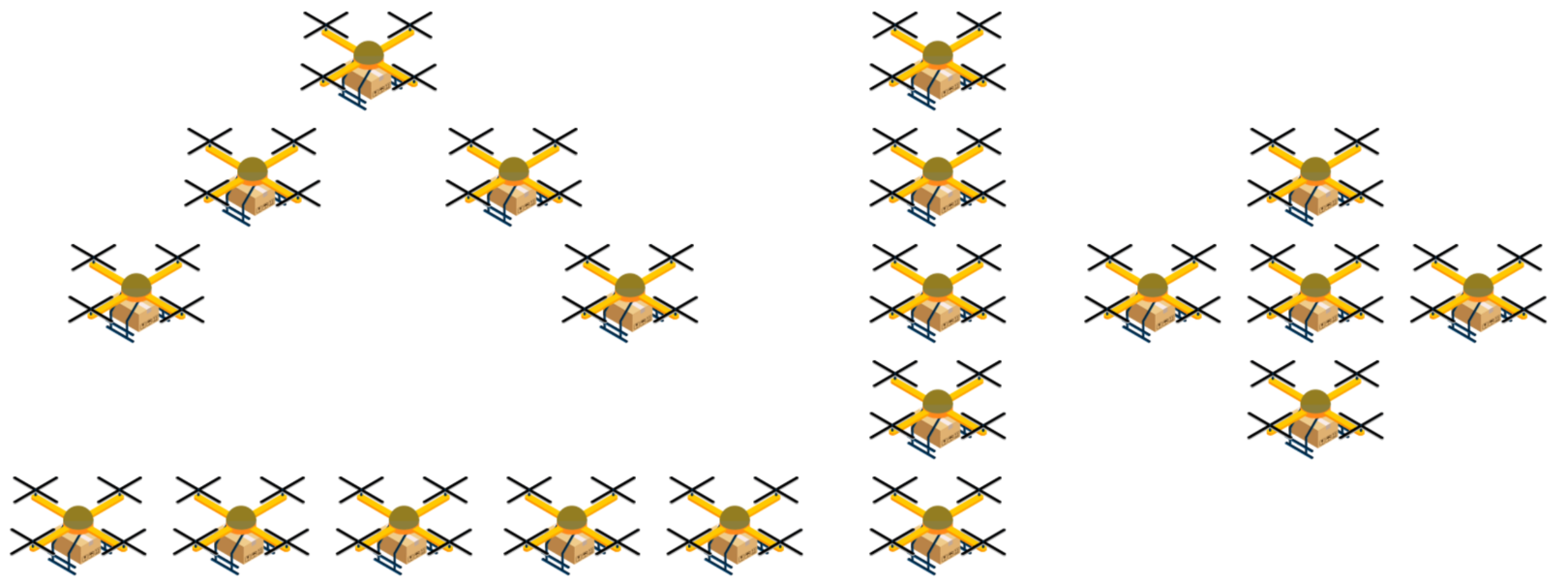}
    \caption{Different Flight Formations by Drone Swarms}
    \label{formations}
\end{figure}

Drone swarms exhibit characteristics that facilitate their delivery operation. First, drone swarms behaviour may be classified into static and dynamic. A static swarm is a swarm whose members are formed at the source node and travel together as a single unit to the destination. A dynamic swarm, in contrast, may split and merge at intermediate nodes between the source and the destination \cite{akram2017security}. Therefore, static swarms would typically compose services (skyway segments) sequentially whereas dynamic swarms may compose multiple services in parallel reducing congestion at intermediate nodes \cite{alkouz2020swarm}. Second, a drone swarm may travel under different fly formations. A swarm may be shaped in a vee, line, column, diamond, etc. (Fig.\ref{formations}). Formation flying assists drones in conserving energy and increases their flight range. This is due to the reduced drag forces from drones blocking wind and the upwash/downwash forces generated by neighbouring drones that provide lift forces \cite{alkouz2020formation}. Third, swarms may move in different ways ensuring the drones stay close safely. This includes, stigmergy, random, and flocking movements adhering to rules of alignment, separation, and cohesion \cite{avvenuti2018detection}.

\subsection{Service-Oriented Architecture of Sensor-Cloud Infrastructure}

\begin{figure} [htp!]
    \centering
    \includegraphics[width=0.9\linewidth]{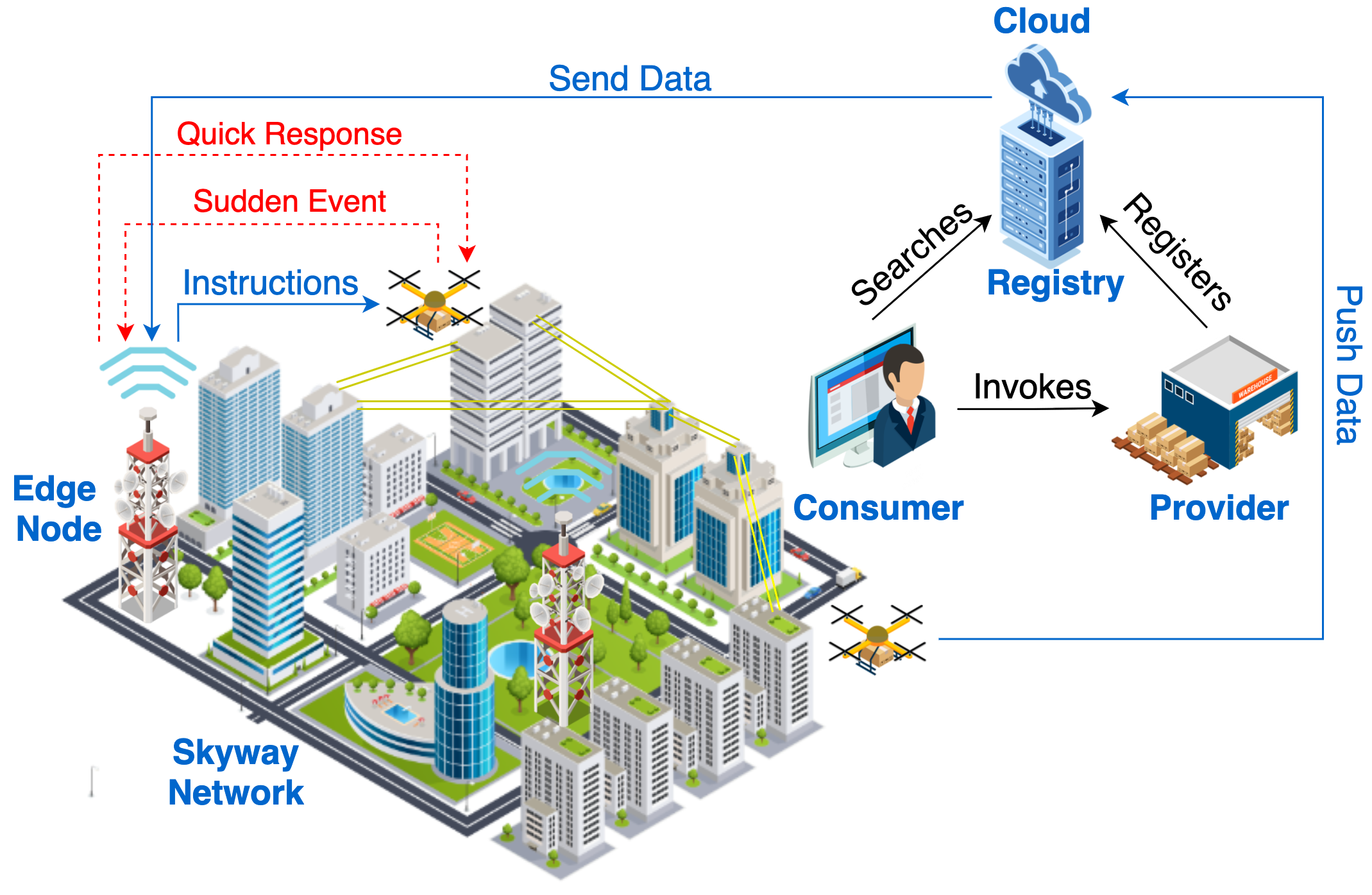}
    \caption{Service-oriented Drone Delivery Architecture}
    \label{Architecture}
\end{figure}

A service-based drone delivery architecture is premised on the skyway network deployment (Fig.\ref{Architecture}). The architecture is made up of two primary parts: a \textit{service-oriented} framework at its core and a \textit{sensor-cloud} infrastructure that serves as a technology enabler. We assume that the components of the architecture interact with each other through a hybrid approach of \textit{orchestration} and \textit{choreography}.

Service providers, service consumers, and service registries are among the components of the service-oriented framework. The communication between services and clients is achieved with a message-based interaction protocol \cite{alshinina2017performance}. Service orientation allows components to be loosely coupled, allowing some components to provide services while others consume those services \cite{jana2006service}. As illustrated in Fig.\ref{Architecture}, drone delivery service \textit{providers} register themselves and their services in the service registry. The \textit{registry} is a database that is updated on a regular basis with information regarding various drone delivery services. The consumers use the service registry to search and invoke providers for required services. When a consumer invokes a service, the drone delivery management system assigns the delivery drone and decides the path to be travelled. The delivery management system makes use of a sensor-cloud infrastructure to provide low-power drones with server-level processing capability.

Drone-based delivery services are a sophisticated technology that necessitates a lot of computing power. Drones, on the other hand, are low-powered, lightweight, and compact in size to reduce take-off weight and maximise flight time \cite{genc2017flying}. Therefore, we use a sensor-cloud architecture to shift processing workloads from drones to edge nodes and the cloud. Simple computations, such as battery and sensor data collection, are performed at the \textit{drone} level. The \textit{edge} nodes are used to offload more computationally intensive jobs. The delivery services path composition is one of these jobs. Edge nodes are often placed at strategic locations throughout the city to reduce latency and improve overall response times. The \textit{cloud} will be used for computations that require a vast amount of data, such as skyway navigation. As a drone travels across the network, it sends its data to the cloud (such as battery level and location). The cloud delivers this data to the edge nodes on a regular basis to aid in speedy decision-making. Multiple instructions are sent from the cloud to the drone via the edge server. The path to the destination, recharging stops, and package pick-up and drop-off commands are all included in these instructions. In the event of a failure and the necessity for a speedy decision, a drone communicates its data to the edge, which makes the decision directly, as shown in Fig. \ref{Architecture}. These decision are later reported back to the cloud. 

\begin{figure} [htp!]
    \centering
    \includegraphics[width=0.9\linewidth]{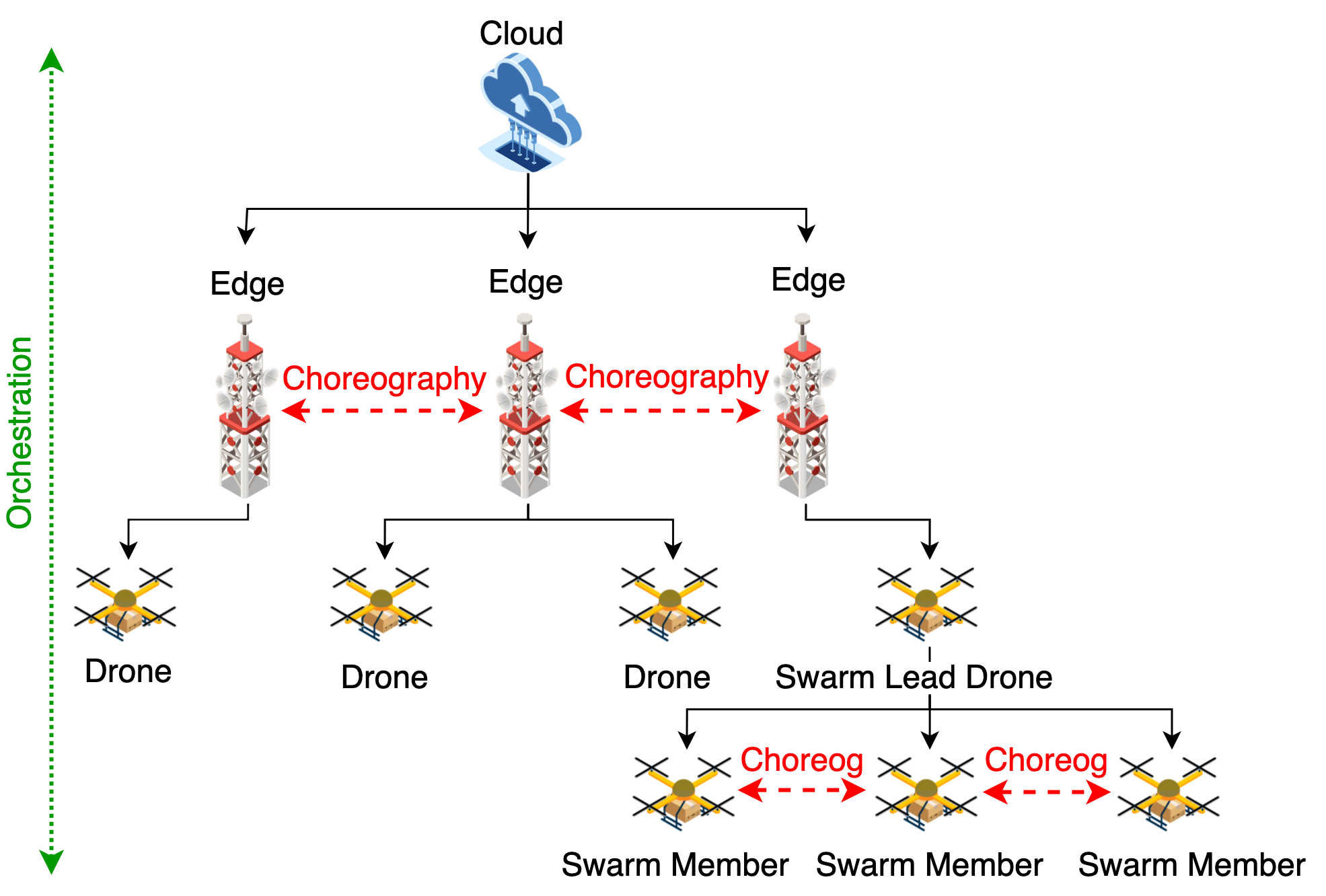}
    \caption{Orchestration and Choreography between Architecture Components}
    \label{orchestration}
\end{figure}

In a service-oriented architecture, \textit{orchestration} refers to the component that manages all elements and interactions between components under its control \cite{cherrier2018services}. In a service-based delivery, the cloud typically manages the transactions between the edge nodes. The cloud would typically send relevant service information to specific concerned edge nodes within the service coverage area. An edge node works in a similar way, orchestrating drones/swarms inside its service area. As shown in Fig.\ref{orchestration}, orchestration is defined as the vertical interaction between the architecture components. The orchestration is viewed as the abstraction or static composition level where a centralized decision may be made. On the contrary, \textit{choreography} is the dynamic execution and response to a sudden occurring event. In this case, a decentralized approach for service composition is necessary. Therefore, in a case of failure for example, the edge nodes dealing with the concerned services would choreograph to re-compose the path quickly before referring back to the cloud. Similarly, within a swarm, if the lead faces problems, the swarm members would choreograph with each other without requiring supervision and instructions from the lead. Therefore, the choreography within a service-based drone delivery is the horizontal dynamic decentralized interactions. This \textit{hybrid} approach blends orchestration and choreography to combine their respective strengths. \textit{The orchestration provides better visibility and has better control whereas choreography has more reactive bearing} \cite{singhal2019selection}.





\section{Contemporary Approaches for Service-Based Drone Delivery}
In this section, we discuss the present approaches for service-based drone delivery in terms of path composition, aerial highways design, and Unmanned Traffic Management (UTM) systems.

\subsection{Drone Service Composition}
Service composition techniques focus on finding an effective combination of atomic services \cite{ Bouguettaya:2017:SCM:3069398.2983528}. Drone service composition approaches ensure congruent and effective provisioning of drone-based deliveries. In this regard, the composition of drone services is defined as the combination of various skyway segments served by a drone or a swarm of drones for the successful delivery of packages to the destination \cite{shahzaad2019composing,alkouz2020swarm}. In constrained multi-drone skyway networks, the drone service composition involves the selection of the optimal recharging stations in addition to the selection of best drone services for the delivery of packages \cite{shahzaad2019constraint}. The service composition approaches based on single drone and swarm of drones pose a set of common challenges to address for the realization of effective delivery services. The common challenges during the composition process include the inherent drone limitations and environmental uncertainties. In addition to common challenges, compositions based on swarm of drones have their specific challenges to address. An example of a specific challenge is the formation of a drone swarm to serve a delivery request.

There is an increasing number of studies that focus on drone service compositions using a single drone and a swarm of drones. A heuristic-based drone service composition algorithm is proposed to select and compose the neighbour drone services taking into account the QoS properties \cite{shahzaad2019composing}. In this paper, the heuristic is based on a spatio-temporal A* (A-star) search algorithm. The objective of this study is to minimize the delivery time and cost for drone-based services. A game-theoretic approach is proposed to compose drone delivery services considering recharging constraints \cite{9284115}. The proposed approach is based on a non-cooperative game algorithm for the selection and composition of optimal drone delivery services. This study aims at minimizing the delivery time and computational time to compute an optimal drone service composition. A swarm-based drone service composition framework is proposed for package delivery in a skyway network \cite{alkouz2020swarm}. The proposed framework includes sequential services composition and parallel services composition algorithms. A cooperative behavior model is also introduced to improve the overall delivery time of swarm-based drone services. A re-allocation framework is proposed for the allocation and scheduling of swarm-based drone services \cite{alkouz2021reinforcemnt}. The proposed framework is composed of two main modules. The first module performs the composition of swarm-based delivery services. The second module uses the output of the first module to allocate and re-allocate the provider owned drones. The objective of the re-allocation is to maximize the provider's profit and drones' utilization.

\subsection{Aerial Highways}
\begin{figure} [h]
    \centering
    \includegraphics[width=0.8\linewidth]{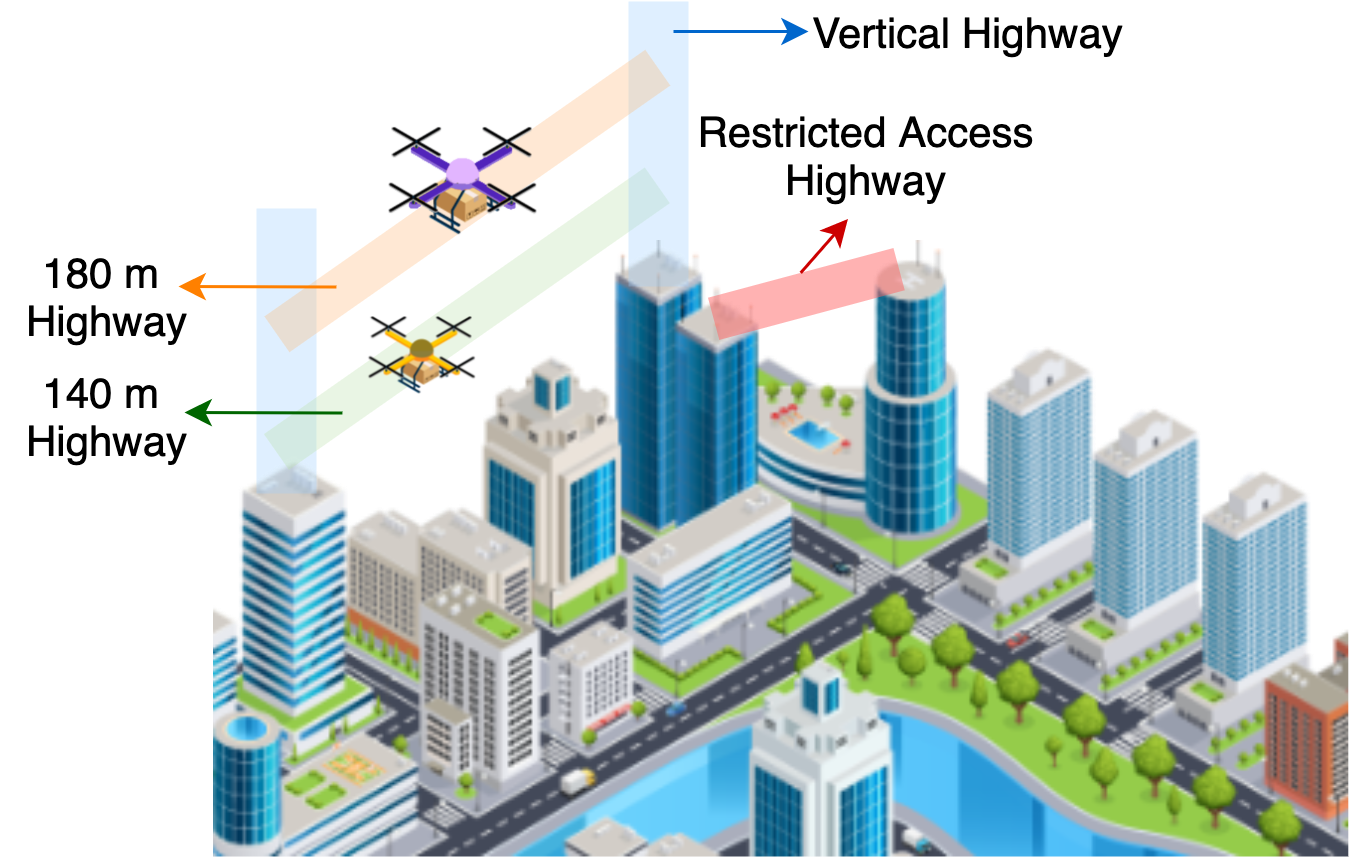}
    \caption{3D Aerial Highways}
    \label{highway}
\end{figure}

As the market for drone package delivery expands, delivery drones will require smooth and dependable cellular communication. Thus, structured and connected sky routes are required. 3D airspace routes, similar to ground vehicle roads, should be established to safely and efficiently carry out delivery drone operations \cite{cherif20213d}. These routes are intentionally planned at various altitudes based on predefined parameters (e.g., drone type, properties, and payload). Vertical pathways are also designed for smooth transitions and uninterrupted cellular coverage. The drones may be guided to fly in different routes based on the cargo priority level. A premium shipment, for example, can be delivered in less time by taking more direct priority routes. In a similar manner, urgent cargo may be given access to restricted  aerial routes (No-Fly Zone). In addition, delivery drones would use the safest aerial routes in case of confidential cargo (e.g., official documents) to guarantee mission integrity \cite{cherif20213d}. Fig. \ref{highway} illustrates a 3D aerial highway vision with drones operating at different altitudes.

\subsection{Unmanned Traffic Management (UTM)}

Drone technology is fast evolving to address pressing issues in transportation, commerce, and public service. The use of technology to plan and manage drone flights will have a significant impact on the airspace's safety and security \cite{kopardekar2016unmanned}. It will also ensure that the airspace is used efficiently, opening up new and exciting business opportunities. \textit{Unmanned Traffic Management} (UTM) is a term that refers to a collection of technologies, services, and processes that will be developed to manage drone operations in and around cities \cite{kopardekar2014unmanned}. Complementary to Air Traffic Management (ATM) systems, a UTM is an ecosystem for uncontrolled drones operations \cite{sandor2019challenges}. Traditional ATM aviation relies on manual technology and processes for air traffic control, such as voice-based communication, individual flight controllers, and take-off and landing clearances. This ATM model cannot support the volume and diversity of expected drones operations. Therefore, a UTM will enable multiple drone operations to be conducted beyond visual line-of-sight (BVLOS), where air traffic services are not provided because of the low altitude operations of drones \cite{kopardekar2016unmanned}.

\begin{figure} [h]
    \centering
    \includegraphics[width=0.8\linewidth]{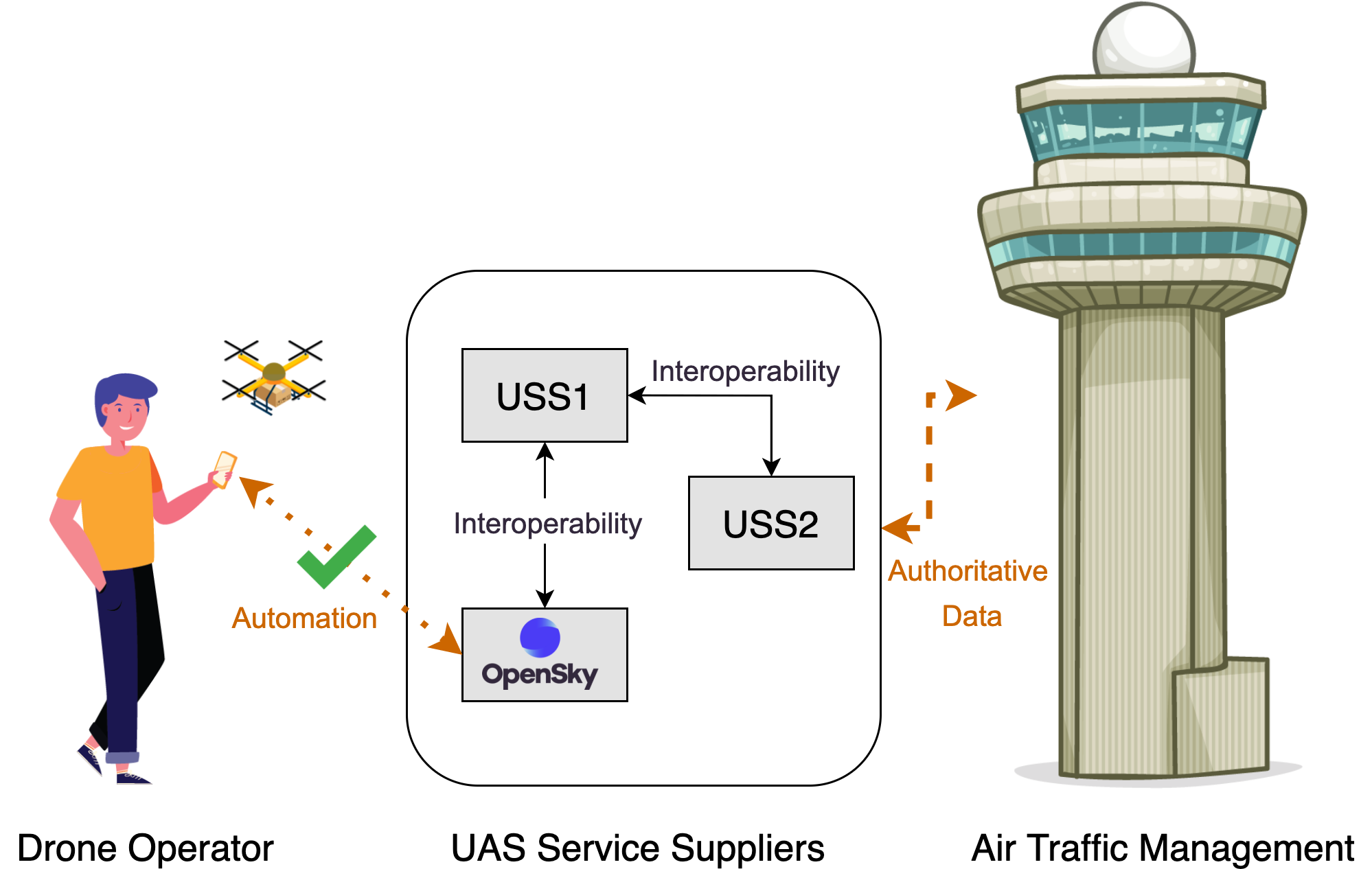}
    \caption{Unmanned Traffic Management Ecosystem}
    \label{utm}
\end{figure}

The FAA, NASA, and industry have collaborated to establish a digital model for UTM and are exploring principles of operation, data exchange needs, and a supporting framework to allow various drone operations\footnote{https://www.faa.gov/uas/research\_development/traffic\_management/}. The UTM ecosystem composes of multiple Unmanned areial systems Service Suppliers (USS). Each USS helps drone operators to plan their flights safely. OpenSky\footnote{https://opensky.wing.com/} is Wing's USS that allows operators to plan their flights and get approval to fly in a certain region at a certain time. The USSs' communicate with each other to avoid flight conflicts and ensure a fare share access to the airspace. A USS may also communicate with aviation authorities and ATMs to facilitate access to restricted air space \cite{kopardekar2016unmanned}. Fig.\ref{utm} illustrates the UTM ecosystem components communications.

\section{Open Challenges and Visioned Directions}


We discuss open challenges in service-based drone delivery and propose a promising new research roadmap. The roadmap focuses on five emerging research areas, namely  drone technologies, skyway infrastructure, service composition, traffic management, and public attitude.

\begin{itemize}
    \item \textbf{Drone Technologies:} 
    As the need for drone delivery grows, serious challenges relating to the primary technology enabler, i.e. drones, must be addressed. One of the major concerns is the \textit{safe dropping of packages} to their final destinations. The two main observed methods currently include drone landing and lowering packages while in air. However, the first option may result in the drone being tampered with. The second method is not feasible with strong wind conditions specially if the package is fragile or liquid. Therefore, more research in this direction needs to be done to ensure the safe landing of packages. Furthermore, \textit{collision and obstacles avoidance} techniques need to be developed, particularly in the case of unanticipated obstacles like birds. Collaborative research with ornithologists is essential after reported bird attacks incidents\footnote{https://www.businessinsider.com.au/watch-australias-google-delivery-drone-attacked-by-raven-mid-air-2021-9}. Last, the area of \textit{noise reduction} is important for a mass adoption of drone deliveries. Current drones are extremely loud. Having a large number of drones flying around cities may result in noise pollution. The use of directional noise, which directs all the noise upwards, is one open area for noise reduction \cite{wu2019novel}.
    \item \textbf{Skyway Infrastructure:}
    Planning the skyway operating environment is essential for successful quick deliveries using drones. However, there is a major lack of research in this area. For example, the optimal positioning and \textit{distribution of recharging pads} across the network may be studied. This is essential to reduce waiting times due to sequential charging caused by congestion. Furthermore, multiple \textit{skyway network topologies} might be researched to meet the needs of different cities. For example, a dense city may require a different structure and distribution of nodes from sparse widely spread cities. These infrastructure plans should consider the optimal \textit{positioning of edge servers} across the network and the \textit{planning of 3D aerial highways} to eliminate accidents.
    \item \textbf{Service Composition:}
   Current research on the composition of drone delivery services concentrate mostly on the deterministic planning of drone delivery services \cite{shahzaad2019composing}. Deterministic planning presupposes that the operating environment remains constant and focuses solely on pre-flight optimization. Therefore, the area of service composition during \textit{uncertainties} is critical. Uncertainties could be caused by sudden weather changes or recharging pads unavailability. Furthermore, in the case of \textit{service failure}, robust approaches need to be studied for an efficient and safe recovery. Moreover, service composition involving the delivery of\textit{ multi-packages} in a single trip could be explored. In another context, more studies are needed on \textit{swarm-based delivery} services. For example, the study of \textit{flight formations} and the effect of \textit{drones re-ordering} within a swarm could be further investigated. In addition, the effect of \textit{homogeneity and heterogeneity of swarms} on the operations is an open research area. Moreover, the possibility of \textit{sharing energy wirelessly} between swarm members while in-flight to increase the flight range could be studied \cite{lakhdari2020crowdsharing}. Finally, the optimal allocation of drone swarms to service request should be explored more \cite{alkouz2021reinforcemnt}.


    
    \item \textbf{Traffic Management:}
    Current UTM regulations focus mainly on the safe operation of multiple drone services. Although safety is very essential, the research on management could be extended to deal with the \textit{efficient} management of drone services. The future research could investigate \textit{robust and stable control algorithms} that enable multiple drones to coordinate their motion. In addition, flight approval \textit{automation} and the representation of \textit{air traffic flow patterns} could potentially be researched. 


    \item \textbf{Public Attitude:} The main perceived risks related to the negative attitude toward drone delivery include performance malfunction and privacy \cite{yoo2018drone}. Therefore, for a real mass adoption of drone delivery, a focused research on eliminating causes of fears should be conducted. For example, research on \textit{safe landing and take-off, designated drop-off areas for privacy concerns, noise reduction, and new technology acceptance psychology} must be conducted.

\end{itemize}







\section{Conclusion}
We presented a novel paradigm for service-based drone delivery to utilize a multi-route skyway infrastructure in drone deliveries. We highlighted the benefits of service-based drone deliveries from service providers and consumers perspectives. We then proposed a service-oriented framework that utilizes a sensor-cloud infrastructure for optimal drones operations in a smart city. Finally, we reviewed current solutions and open challenges, as well as future visioned directions for research in service-based drone deliveries. 

\section*{Acknowledgment}
This research was partly made possible by DP160103595 and LE180100158 grants from the Australian Research Council. The statements made herein are solely the responsibility of the authors.

\bibliographystyle{IEEEtran}  
\bibliography{CIC}

\end{document}